\documentclass[runningheads]{llncs}
\usepackage[T1]{fontenc}
\usepackage{graphicx}
\usepackage{siunitx}   
\DeclareSIUnit{\litre}{l}
\usepackage{subcaption}
\usepackage{multirow}
\usepackage{tabularx}
\usepackage{makecell}
\usepackage[table,xcdraw]{xcolor}
\usepackage{color}
\usepackage{float}

\begin{document}
\title{Influence of Water Droplet Contamination for Transparency Segmentation}
%
%
\author{Volker Knauthe\inst{1}\orcidID{0000-0001-6993-5099} \and
 Paul Weitz \inst{1}\orcidID{0009-0000-6032-0637} \and
 Thomas P\"ollabauer\inst{2,1}\orcidID{0000-0003-0075-1181} \and
 Tristan Wirth\inst{1}\orcidID{0000-0002-2445-9081} \and
 Arne Rak\inst{1}\orcidID{0000-0001-6385-3455} \and
 Arjan Kuijper\inst{2}\orcidID{0000-0002-6413-0061} \and
 Dieter\,W. Fellner\inst{1,2,3}\orcidID{0000-0001-7756-0901}}

\authorrunning{V. Knauthe et al.}
\titlerunning{Influence of Water Droplet Contamination for Transparency Segmentation}
 
%
 \institute{Technical University of Darmstadt, Darmstadt, Germany \and
 Fraunhofer Institute for Computer Graphics Research IGD, Darmstadt, Germany \and
 CGV Institute, Graz University of Technology, Graz, Austria}
\maketitle              
\begin{abstract}
Computer vision techniques are on the rise for industrial applications, like process supervision and autonomous agents, e.g., in the healthcare domain and dangerous environments.
While the general usability of these techniques is high, there are still challenging real-world use-cases.
Especially transparent structures, which can appear in the form of glass doors, protective casings or everyday objects like glasses, pose a challenge for computer vision methods.
This paper evaluates the combination of transparent objects in conjunction with (naturally occurring) contamination through environmental effects like hazing.
We introduce a novel publicly available dataset containing 489 images incorporating three grades of water droplet contamination on transparent structures and examine the resulting influence on transparency handling.
Our findings show, that contaminated transparent objects are easier to segment and that we are able to distinguish between different severity levels of contamination with a current state-of-the art machine-learning model.
This in turn opens up the possibility to enhance computer vision systems regarding resilience against, e.g., datashifts through contaminated protection casings or implement an automated cleaning alert.

\end{abstract}
\section{Introduction}
The supervision of industrial processes and the usage of autonomous agents in everyday live becomes more and more prevalent in our world. Due to the unpredictable nature of the majority of real-world tasks, this naturally leads to uncontrolled environmental conditions. This in turn opens up a variety of new challenges for machines to properly interact with the surrounding world. Our work focuses on one of those aspects, namely the detection of progressively contaminated transparent objects. While transparent objects already pose a challenge themselves, they undergo a faster and more drastic change of appearance due to contamination than opaque objects. In turn, this appearance change influences two major interactive properties of a transparent object. First, the detection of said objects is affected due to a shift in material visibility. Second, the ability to correctly recognize objects behind an increasingly contaminated transparent surface becomes more challenging up to a point of being impossible. This is especially relevant for, e.g., vision systems (monitoring production/chemistry prone to contamination) that are behind a transparent safety glass or autonomous agents in transparency affine environments like hospitals. It is therefore of interest to know the grade of transparency contamination to assess the quality and reliability of an original intended vision task.\\
To provide more insights about the mentioned challenges, we introduce a novel real world dataset, which consists of 489 images with three degrees of contamination. With this dataset we perform two major experiments utilizing the \textit{Trans10K} dataset \cite{xie2020segmenting} for additional data and \textit{Trans4Trans} \cite{zhang2021trans4trans} as base model. Our findings emphasize, that the contamination of transparent objects makes them easier to detect and that the severity of contamination is distinguishable. With this insight, it is possible to detect detrimental inference for vision tasks that look through transparency and, e.g., call for cleaning assistance or further assess anomalies or wrong predictions from the original task.

\section{Related Work}

In this chapter we discuss recent advancements and state-of-the-art strategies dealing with semantic segmentation (see \ref{sec:related:semantic}) and transparency segmentation (see \ref{sec:related:transparency}). We further give an overview over the relevant work that examines contamination on transparent and opaque surfaces (see \ref{sec:related:contamination}), showing that the influence on transparent structures regarding the task of transparency segmentation and contamination severity estimation has not been addressed in the literature yet.

\subsection{Semantic Segmentation}
\label{sec:related:semantic}

Semantic Segmentation describes the task of assigning separate class labels to each pixel of an input 2D image \cite{ulku2022survey}.
Early publications leverage convolutional neural networks (CNNs) \cite{chen2014semantic,long2015fully,shelhamer2017fully}.
Several authors \cite{chen2017deeplab,lin2016efficient,zheng2015conditional} propose the usage of conditional random fields to improve the segmentation results especially in the area of object boundaries.
In general, encoder-decoder based architectures \cite{badrinarayanan2017segnet,iglovikov2018ternausnetv2,ronneberger2015u} exhibit high segmentation performance. 
The adoption of architectural designs, such as feature pyramid pooling \cite{chen2017deeplab,chen2018encoder,zhao2017pyramid} or spatial pyramid pooling \cite{he2015spatial,li2018pyramid} have further improved the quality of estimated segmentations.\\
Recently, attention strategies have been adapted from the domain of natural language processing \cite{vaswani2017attention} into the domain of computer vision \cite{dosovitskiy2020image}.
Attention models the dependencies of sequence elements, i.e., image patches, without regard to their distance in the input or output sequence \cite{vaswani2017attention}.
Multiple variations of the attention mechanism have further improved the state-of-the-art performance of segmentation models \cite{thisanke2023semantic}.
SETR \cite{zheng2021rethinking} and Segmenter \cite{strudel2021segmenter} use end-to-end transformer architectures.
Picking up the idea of pyramid architectures, Segformer \cite{xie2021segformer} and Pyramid Vision Transformer (PVT) \cite{wang2021pyramid,wang2022pvt} employ hierarchical transformer architectures.
Chu et al. \cite{chu2021twins} mitigate the limitation of PVT to fixed input size by incorporating Conditional Position encoding Vision Transformer (CPVT) \cite{chu2021conditional}.
Yuan et al. \cite{yuan2021hrformer} propose High-Resolution Transformer that enable predictions on high resolution images using multi-resolution parallel transformer.
Swin Transformer \cite{liu2021swin} utilize a shifted window attention mechanism reducing the computational complexity of the attention mechanism from quadratic to linear.
Masked-attention Transformer \cite{cheng2022masked} limit the relevant regions for cross-attention to the image foreground, reducing complexity even further.
Some contributions \cite{cheng2022masked,jain2023oneformer} enhance the performance on semantic segmentation by formulating a general segmentation (instance, semantic, panoptic) as a multi-task training problem.\\
In contrast to that, InternImage-H \cite{wang2023internimage} shows impressive semantic segmentation results with CNN-based vision transformer with deformable convolutions, enhancing their receptive field, effectively mitigating the drawbacks of CNNs in comparison to transformer models.
Su et al. \cite{su2023towards} further improve InternImage-H by integrating an all-in-one single-stage pre-training approach.\\
Recently, foundation models, such as EVA \cite{fang2023eva}, DinoV2 \cite{oquab2023dinov2} and SAM \cite{kirillov2023segment}, that leverage vast amounts of training data, have further improved the state-of-the-art performance on a multitude of vision tasks including semantic segmentation.
Bringing foundation models even further, recent models such as BeiT-3 \cite{wang2022image} and ONE-PEACE \cite{wang2023one} incorporate multi-model data including audio and language leading to even better results on semantic segmentation.

\subsection{Transparency Segmentation}
\label{sec:related:transparency}

Transparency Segmentation is a mode of semantic segmentation, where either transparent structures are discriminated against other structures or more refined classes of transparent objects are labeled on a pixel base, e.g, Trans10K \cite{xie2020segmenting,xie2021segmenting}.\\
Some strategies leverage supplementary information in addition to image inputs for transparency segmentation. 
Transcut \cite{xu2015transcut} bases its estimations on a light field.
Tom-Net \cite{chen2018tom} requires a refractive flow map as label during training, which is hard to obtain from the real world.
Huo et al. \cite{huo2023glass} incorporate thermal image data into their segmentation process.
However, in the context of this work, we consider strategies that utilize additional information out of scope.\\
In constrast, a multitude of architectures only require the input of a single RGB-image.
TransLab \cite{xie2020segmenting} utilises ResNet \cite{he2016deep} as the backbone network and incorporates boundary prediction to improve transparency detection by focusing on the contrasting edges of transparent objects.
Trans2Seg \cite{xie2021segmenting} employs a hybrid CNN-transformer-based segmentation pipeline consisting of a CNN backbone for feature extraction and a transformer encoder and decoder.
Zhang et al. \cite{zhang2021trans4trans,zhang2022trans4trans} propose Trans4Trans, that is considered the current state-of-the-art.
Trans4Trans utilizes Pyramid Vision Transformer \cite{wang2021pyramid} in the encoder stage combined with a transformer-based decoder.
The authors claim that the transformer-based decoder stage makes the model more resilient against unseen data.\\
Knauthe et al. \cite{knauthe2023distortion} conducted a perception study and trained a neural network, that emphasize the correlation between human/machine perception capabilities and the strength of image distortions effects. However, their research utilizes a synthetic dataset, that simulates varying global distortions on panorama image crops. Therefore their work is not applicable to our contribution, due to the requirement of localized gradually contaminated transparent objects in the wild for our novel contamination related segmentation task.

\subsection{Surface Contamination}
\label{sec:related:contamination}
Some work addresses the detection of dirt contamination on opaque objects like the floor \cite{bormann2020dirtnet,canedo2021deep}, solar panels \cite{olorunfemi2022solar}, wind turbines \cite{jimenez2019dirt} and conveyor belts \cite{canedo2021deep}, to optimize cleaning tasks.
Furthermore, the detection of soiling \cite{shajahan2021camera,uricar2021let} and damage \cite{nguyen2022detection} to camera lenses has been discussed in recent work.
Some authors tackle the mitigation of the effects of water droplets for images captured through a windshield \cite{eigen2013restoring,halimeh2009raindrop,quan2019deep}.
To our knowledge the influence and severity estimation of contamination on transparent surfaces for semantic segmentation, which are discussed in this paper, have not been examined up to this point.

\section{Real-World Transparency Contamination Dataset}

\begin{figure}[p]
\centering
\begin{subfigure}{1.0\textwidth}
\begin{subfigure}{.24\textwidth}
  \centering
  \includegraphics[width=1.0\linewidth]{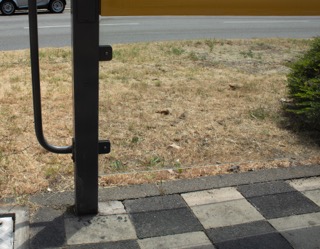}
\end{subfigure}
\hfill
\begin{subfigure}{.24\textwidth}
  \centering
  \includegraphics[width=1.0\linewidth]{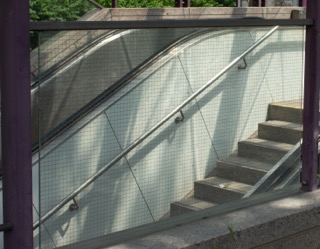}
\end{subfigure}
\hfill
\begin{subfigure}{.24\textwidth}
  \centering
  \includegraphics[width=1.0\linewidth]{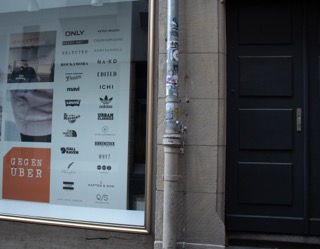}
\end{subfigure}
\hfill
\begin{subfigure}{.24\textwidth}
  \centering
  \includegraphics[width=1.0\linewidth]{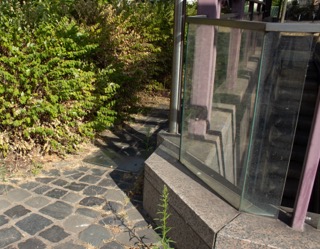}
\end{subfigure}
\caption{Examples of images depicting a single instance of a transparent item in an object-centric perspective with no obstruction.}
\label{fig:single_objects}
\end{subfigure}

\begin{subfigure}{1.0\textwidth}

\begin{subfigure}{.24\textwidth}
  \centering
  \includegraphics[width=1.0\linewidth]{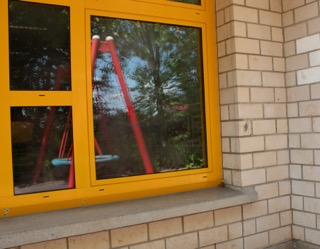}
\end{subfigure}
\hfill
\begin{subfigure}{.24\textwidth}
  \centering
  \includegraphics[width=1.0\linewidth]{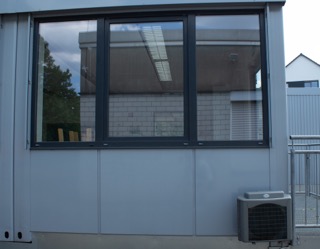}
\end{subfigure}
\hfill
\begin{subfigure}{.24\textwidth}
  \centering
  \includegraphics[width=1.0\linewidth]{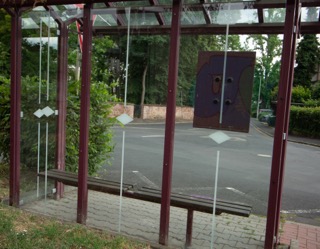}
\end{subfigure}
\hfill
\begin{subfigure}{.24\textwidth}
  \centering
  \includegraphics[width=1.0\linewidth]{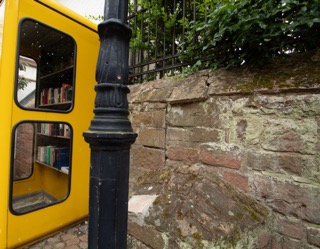}
\end{subfigure}
\caption{Examples of images depicting multiple transparent objects of similar appearance.}
\label{fig:multiple_objects}
\end{subfigure}

\begin{subfigure}{1.0\textwidth}
\begin{subfigure}{.24\textwidth}
  \centering
  \includegraphics[width=1.0\linewidth]{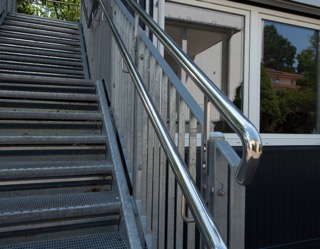}
\end{subfigure}
\hfill
\begin{subfigure}{.24\textwidth}
  \centering
  \includegraphics[width=1.0\linewidth]{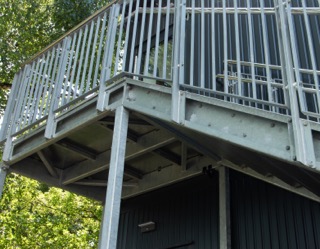}
\end{subfigure}
\hfill
\begin{subfigure}{.24\textwidth}
  \centering
  \includegraphics[width=1.0\linewidth]{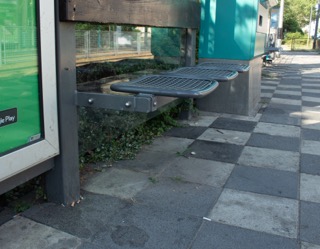}
\end{subfigure}
\hfill
\begin{subfigure}{.24\textwidth}
  \centering
  \includegraphics[width=1.0\linewidth]{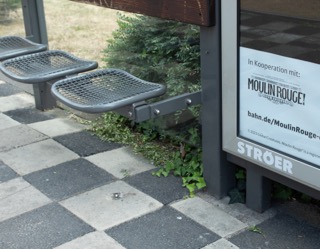}
\end{subfigure}
\caption{Examples of images depicting occluded transparent objects.}
\label{fig:occluded_objects}
\end{subfigure}

\begin{subfigure}{1.0\textwidth}
\begin{subfigure}{.24\textwidth}
  \centering
  \includegraphics[width=1.0\linewidth]{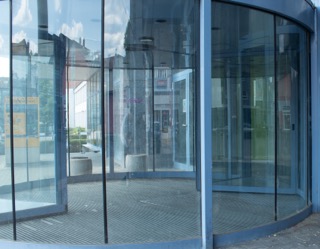}
\end{subfigure}
\hfill
\begin{subfigure}{.24\textwidth}
  \centering
  \includegraphics[width=1.0\linewidth]{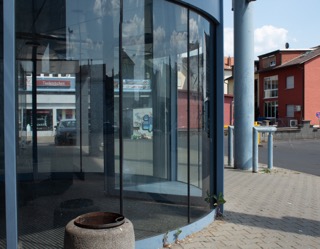}
\end{subfigure}
\hfill
\begin{subfigure}{.24\textwidth}
  \centering
  \includegraphics[width=1.0\linewidth]{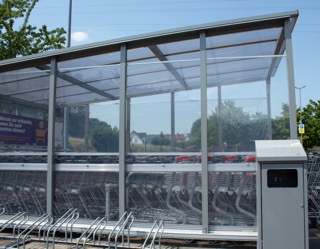}
\end{subfigure}
\hfill
\begin{subfigure}{.24\textwidth}
  \centering
  \includegraphics[width=1.0\linewidth]{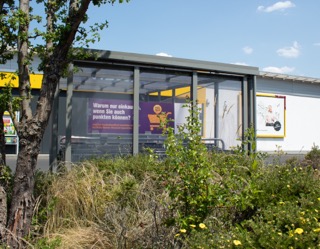}
\end{subfigure}
\caption{Examples of images depicting complex transparent objects with multiple layers.}
\label{fig:complex_objects}
\end{subfigure}

\begin{subfigure}{1.0\textwidth}

\begin{subfigure}{.24\textwidth}
  \centering
  \includegraphics[width=1.0\linewidth]{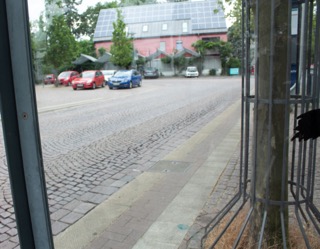}

\end{subfigure}
\hfill
\begin{subfigure}{.24\textwidth}
  \centering
  \includegraphics[width=1.0\linewidth]{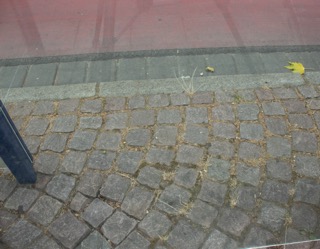}
\end{subfigure}
\hfill
\begin{subfigure}{.24\textwidth}
  \centering
  \includegraphics[width=1.0\linewidth]{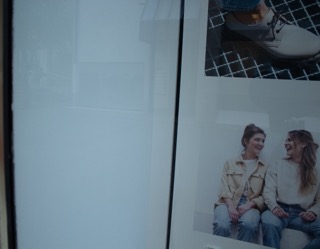}
\end{subfigure}
\hfill
\begin{subfigure}{.24\textwidth}
  \centering
  \includegraphics[width=1.0\linewidth]{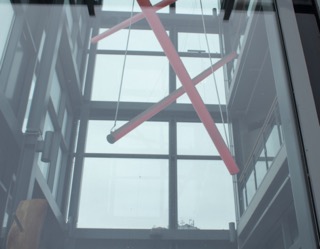}
\end{subfigure}
\caption{Examples of images depicting mostly or only transparent surfaces with little contextual information.}
\label{fig:fully_trans_objects}
\end{subfigure}
\caption{Overview of the different scenes captured for the dataset.}
\label{fig:scene_examples}
\end{figure}

The dataset assembled for this experiment consists of transparent objects that one could encounter in any ordinary urban environment. 
All images were captured using a DSLR camera, utilising various lenses with focal lengths between 10-55mm and apertures ranging from $f/3.5$ to $f/16$, as well as the lowest sensor sensitivity possible to reduce noise to a minimum. 
The scope of the dataset was restricted to daytime scenery to reduce the visual variations of the environments.
While all images were captured in a downtown setting, the actual objects and their appearances still vary depending on the present surroundings. 
All captured scenes contain one or more transparent surfaces, in cases such as several windows of the same type.
Transparent objects may be completely exposed, or partially occluded with reflections ranging from basically non-existent to strong environmental reflections, which severely impair the observed transparency.
A selection of scenes captured for the dataset is given in Fig. \ref{fig:scene_examples}.\\
To simulate the presence of contamination, a fine layer of water was applied to each object in two passes. 
For each pass, a uniform density of approximately \SI[per-mode=symbol]{1}{\m\litre\per25\square\cm} of water was applied to the whole surface. 
Before the first and after each subsequent pass, the objects were captured with identical camera settings utilising a tripod, resulting in three images per object. 
In total, the dataset consists of 489 images with three different categories: \textit{no modification}, \textit{1 pass} and \textit{2 passes}.
All objects are labeled on a per-pixel level by ourselves, with the transparent surface being masked by a polygon and assigned the value of the respective contamination class. 
Additionally, pixels belonging to the background were assigned with 0. 
All parts of a transparent surface were labeled, even if parts of the surface appear nontransparent due to the presence of stickers or similar opaque objects.\\
To assure high label quality, the annotation process was audited by two auditors independently. The assigned labels were refined on a per image base until both auditors found them to be precise. The dataset is publicly available via contacting the authors.

\section{Methodology}
\subsubsection{Evaluation Metrics}
To evaluate the differences in segmentation caused by the added contamination, we select three metrics to measure the results of the transparency segmentation model. Precisely, we select the \textit{Pixel Accuracy (PAcc)}, \textit{Category Intersection over Union (eg. tIoU for transparency IoU)} as well as \textit{Mean Intersection over Union (mIoU)}, all of which have been used to evaluate the most recent RGB transparency segmentation models \cite{xie2020segmenting,xie2021segmenting,zhang2021trans4trans}. For the \textit{Intersection over Union}, we specifically focus on the \textit{transparency} class.

%

\subsubsection{Transformer Based Model and SAM}
For the dataset evaluation, we select \textit{Trans4Trans} \cite{zhang2021trans4trans} as the transparency segmentation model. As of the writing of this paper, this model achieves state-of-the-art segmentation results on the \textit{Trans10K} \cite{xie2020segmenting} dataset. 
We choose the \textit{Trans10K} dataset, since the scenes captured for our contamination dataset depict real-world transparent objects in an outdoor environment, which are very similar in context. Therefore, the use of a well-performing model regarding such data is logical.
In addition to the \textit{Trans4Trans} model, we also evaluate the 5 test sets on a foundation model, namely \textit{Segment Anything Model (SAM)} \cite{kirillov2023segment}. To achieve this, we let SAM segment each image present in the test set of a given split. Because \textit{SAM} segments the whole image, we calculated the \textit{IoU} values for every segment detected by \textit{SAM}. We then reported the highest \textit{IoU} value for the observed image. After repeating this process for the whole test set, we averaged the results and reported this value for the respective split.

\subsubsection{Training Process}
\label{sec:training_process}
The process for training and evaluating the \textit{Trans4Trans} model on our dataset is as follows:\\
(1) We train the model on a large-scale transparent object dataset. For this, we select the \textit{Trans10K} dataset, as it features a large quantity of images depicting transparent objects with dense annotations for transparency segmentation. Since the main goal of this experiment is to observe the detection performance of a model in relation to the amount of contamination present on a surface, the categories of the \textit{Trans10K} dataset are reduced to the classes \textit{background} and \textit{transparency} during the data loading process. This puts the focus of the experiment on detecting transparency itself, rather than detecting different types of objects. For the pretrained backbone of the \textit{Trans4Trans} architecture, the \textit{PVT-Medium} model is chosen, as it achieves the best performance on the \textit{Trans10K} dataset. The network is trained on four Nvidia A100 SXM4 GPUs for 100 epochs with a batch size of 4 images per iteration for each GPU. All images are cropped to (512, 512) during data preparation. The learning rate is initialised with $1*10^{-4}$ and scheduled utilising poly strategy \cite{yu2018bisenet} with a power of 0.9 in 100 epochs. \textsc{ADAMW} is used as the optimiser with epsilon $1e-8$ and weight decay $1e-4$. These values are directly adopted from the training process on the \textit{Trans10K} dataset described in \cite{zhang2021trans4trans} to achieve an outcome as close as possible.\\
(2) We adapt the weights of the trained model to our custom dataset by transfer-training the model. First, we ensure a clean separation between the training, validation and testing split of our dataset by grouping all three different versions of an object together in order to prevent multiple versions of the same object occurring in different splits. Then, random splits of 50\% train, 10\% val and 40\% test are constructed over the different objects, mimicking the distribution of the \textit{Trans10K} dataset. We repeat this process five times to achieve balanced splits in regards to the contained scenery, with a different random seed each time. This allows for averaging the results during the evaluation step and mitigates the impact of uneven scenery distributions that could occur during random splitting. To transfer-train the model, the learning rate and optimiser are reset to their initial values adopted from \cite{zhang2021trans4trans}. To ensure proper adaptation to our data, we do not freeze any weights, which allows the model to fully adapt to any new scenery present in the dataset while still maintaining the overall feature detection learned from the large-scale dataset.
The process is performed on the same hardware as the initial training, with each split having a batch size of 4 images per iteration for each GPU for 4 epochs. This value is chosen because after 4 training epochs, the training loss stagnated at around 0.025, which indicates sufficient fitting to the data. Like before, the images are cropped to (512, 512). \\
In total, five different models are obtained after the training process.
For the evaluation, each model is tested with the test set of its respective split by measuring the segmentation results for all three metrics. To measure the results for each of the contamination classes, the images of interest are filtered out and tested individually.

\section{Results}

\subsection{Effects of Contamination on Transparency Segmentation}
\label{sec:results:contamination_impact}

\subsubsection{Quantitative Results}

The results displayed in Table \ref{results:SAM:quantitative:baseline} show the average \textit{IoU} value of each of the five splits as well as the difference caused by our modification between the contamination classes after testing each test set against the \textit{SAM} model. As can be observed, the application of our simulated contamination did increase the \textit{IoU} value for every split by an average of 7.39 $\%$ between \textit{no contamination} and \textit{little contamination}, and 6.38 $\%$ between \textit{little contamination} and \textit{strong contamination}. This emphasizes that the segmentation quality benefited from the application of our simulated contamination.
\begin{table}[!htbp]
\centering
\setlength{\tabcolsep}{0.1em} 
\renewcommand{\arraystretch}{1.0}
\caption{Results and difference in segmentation quality of the \textit{SAM} model for the 3 classes of each split. $\Delta$ denotes the difference of two adjacent contamination classes for the IoU metric.}
\label{results:SAM:quantitative:baseline}
\begin{tabular}{c|c|c|c|c|c}
\hline
{\textbf{Splits}} & \multicolumn{1}{c|}{no cont. IoU $\uparrow$} & \multicolumn{1}{c|}{$\leftarrow\Delta\rightarrow$} & \multicolumn{1}{c|}{little cont. IoU $\uparrow$} & \multicolumn{1}{c|}{$\leftarrow\Delta\rightarrow$} & strong cont. IoU $\uparrow$ \\ \hline
\textbf{1} & \multicolumn{1}{c|}{32.16} & \multicolumn{1}{c|}{{8.82}} & \multicolumn{1}{c|}{40.97} & \multicolumn{1}{c|}{{5.47}} & 46.45 \\
\textbf{2} & \multicolumn{1}{c|}{32.97} & \multicolumn{1}{c|}{3.84} & \multicolumn{1}{c|}{36.81} & \multicolumn{1}{c|}{7.88} & 44.69 \\
\textbf{3} & \multicolumn{1}{c|}{31.88} & \multicolumn{1}{c|}{4.33} & \multicolumn{1}{c|}{36.22} & \multicolumn{1}{c|}{6.92} & 43.14 \\
\textbf{4} & \multicolumn{1}{c|}{28.61} & \multicolumn{1}{c|}{10.13} & \multicolumn{1}{c|}{38.74} & \multicolumn{1}{c|}{5.69} & 44.43 \\
\textbf{5} & \multicolumn{1}{c|}{28.95} & \multicolumn{1}{c|}{9.82} & \multicolumn{1}{c|}{38.77} & \multicolumn{1}{c|}{5.94} & 44.72 \\ \hline
\multicolumn{1}{l|}{\textbf{Avg.}} & \multicolumn{1}{c|}{\textbf{30.91}} & \multicolumn{1}{c|}{\textbf{7.39}} & \multicolumn{1}{c|}{\textbf{38.30}} & \multicolumn{1}{c|}{\textbf{6.38}} & \textbf{44.68} \\ \hline
\multicolumn{1}{l|}{\textbf{$\sigma$}} & \multicolumn{1}{c|}{\textbf{1.78}} & \multicolumn{1}{c|}{} & \multicolumn{1}{c|}{\textbf{1.68}} & \multicolumn{1}{c|}{} & \textbf{1.05} \\ \hline
\end{tabular}
\end{table}
To gain a better understanding of the general adaptation of the \textit{Trans4Trans} model to our dataset, the segmentation results for objects with no contamination can be observed in Table \ref{table:results:baseline}. The model was able to properly adapt to our data regarding general transparency segmentation.
\begin{table}[H]
\centering
\setlength{\tabcolsep}{0.2em} 
\renewcommand{\arraystretch}{1.1}
\caption{Segmentation performance of the \textit{Trans4Trans} model of the different contamination classes for each split. tIoU denotes the Intersection over Union for the \textit{transparency} class, mIoU denotes the Mean Intersection over Union and PAcc denotes the Pixel Accuracy.}
\label{table:results:baseline}
\begin{tabular}{c|ccc|ccc|ccc}
\hline
                                    & \multicolumn{3}{c|}{}                                                                                    & \multicolumn{3}{c|}{}                                                                                                      & \multicolumn{3}{c}{}                                                                                                      \\
                                    & \multicolumn{3}{c|}{\multirow{-2}{*}{no contamination}}                                                  & \multicolumn{3}{c|}{\multirow{-2}{*}{little contamination}}                                                                & \multicolumn{3}{c}{\multirow{-2}{*}{strong contamination}}                                                                \\ \cline{2-10} 
\multirow{-3}{*}{\textbf{Splits}}   & \multicolumn{1}{c}{tIoU $\uparrow$} & \multicolumn{1}{c}{mIoU $\uparrow$} & PAcc $\uparrow$ & \multicolumn{1}{c}{tIoU $\uparrow$}       & \multicolumn{1}{c}{mIoU $\uparrow$}             & PAcc $\uparrow$ & \multicolumn{1}{c}{tIoU $\uparrow$}       & \multicolumn{1}{c}{mIoU $\uparrow$}             & PAcc $\uparrow$ \\ \hline
\textbf{1}                          & \multicolumn{1}{c}{89.94}         & \multicolumn{1}{c}{92.12}            & 95.56               & \multicolumn{1}{c}{88.71}                        & \multicolumn{1}{c}{91.17}                        & 95.67              & \multicolumn{1}{c}{88.21}                        & \multicolumn{1}{c}{90.76}                        & 95.59              \\
\textbf{2}                          & \multicolumn{1}{c}{89.42}                   & \multicolumn{1}{c}{91.53}             & 94.90               & \multicolumn{1}{c}{{90.30}} & \multicolumn{1}{c}{{92.18}} & 95.64               & \multicolumn{1}{c}{{89.88}} & \multicolumn{1}{c}{{91.80}} & 95.70               \\
\textbf{3}                          & \multicolumn{1}{c}{86.05}                   & \multicolumn{1}{c}{89.09}             & 94.10               & \multicolumn{1}{c}{87.29}                         & \multicolumn{1}{c}{90.08}                         & 94.55               & \multicolumn{1}{c}{89.56}                         & \multicolumn{1}{c}{91.82}                         & 95.36               \\
\textbf{4}                          & \multicolumn{1}{c}{89.55}                   & \multicolumn{1}{c}{91.09}             & 94.78               & \multicolumn{1}{c}{{91.78}} & \multicolumn{1}{c}{{92.97}} & 96.13               & \multicolumn{1}{c}{{91.34}} & \multicolumn{1}{c}{{92.56}} & 96.14               \\
\textbf{5}                          & \multicolumn{1}{c}{87.46}                   & \multicolumn{1}{c}{90.11}             & 94.71               & \multicolumn{1}{c}{88.15}                         & \multicolumn{1}{c}{90.66}                         & 95.35               & \multicolumn{1}{c}{89.07}                         & \multicolumn{1}{c}{91.38}                         & 95.74               \\ \hline
\multicolumn{1}{l|}{\textbf{Avg.}} & \multicolumn{1}{c}{\textbf{88.48}}                   & \multicolumn{1}{c}{\textbf{90.79}}             & \textbf{94.81}               & \multicolumn{1}{c}{\textbf{89.25}}                         & \multicolumn{1}{c}{\textbf{91.41}}                         & \textbf{95.47}               & \multicolumn{1}{c}{\textbf{89.61}}                         & \multicolumn{1}{c}{\textbf{91.66}}                         & \textbf{95.70}               \\ \hline
\multicolumn{1}{l|}{\textbf{$\sigma$}} & \multicolumn{1}{c}{\textbf{1.49}}                   & \multicolumn{1}{c}{\textbf{1.07}}             & \textbf{0.47}               & \multicolumn{1}{c}{\textbf{1.60}}                         & \multicolumn{1}{c}{\textbf{1.04}}                         & \textbf{0.52}               & \multicolumn{1}{c}{\textbf{1.03}}                         & \multicolumn{1}{c}{\textbf{0.59}}                         & \textbf{0.25}               \\ \hline
\end{tabular}
\end{table}
\begin{table}[!htbp]
\centering
\setlength{\tabcolsep}{0.1em} 
\renewcommand{\arraystretch}{1.0}
\caption{Difference in segmentation quality between the 3 contamination classes of each split. For the average of 3 splits, the first and third split have been excluded. $\Delta$ denotes the difference of two adjacent contamination classes for a given metric.}
\label{results:quantitative:diff}
\begin{tabular}{c|ccc|ccc}
\hline
                                    & \multicolumn{3}{c|}{}                                                                                                               & \multicolumn{3}{c}{}                                                                                                                \\
                                    & \multicolumn{3}{c|}{\multirow{-2}{*}{no cont. / little cont.}}                                                                      & \multicolumn{3}{c}{\multirow{-2}{*}{little cont. / strong cont.}}                                                                   \\ \cline{2-7} 
\multirow{-3}{*}{\textbf{Splits}}   & \multicolumn{1}{c}{$\Delta$ tIoU $\uparrow$}                & \multicolumn{1}{c}{$\Delta$ mIoU $\uparrow$}             & $\Delta$ PAcc $\uparrow$ & \multicolumn{1}{c}{$\Delta$ tIoU $\uparrow$}       & \multicolumn{1}{c}{$\Delta$ mIoU $\uparrow$}             & $\Delta$ PAcc $\uparrow$           \\ \hline
\textbf{1}                          & \multicolumn{1}{c}{{-1.23}} & \multicolumn{1}{c}{{-0.95}} & 0.11               & \multicolumn{1}{c}{{-0.50}} & \multicolumn{1}{c}{{-0.41}} & {-0.08} \\ 
\textbf{2}                          & \multicolumn{1}{c}{0.89}                                  & \multicolumn{1}{c}{0.66}                         & 0.74               & \multicolumn{1}{c}{{-0.43}} & \multicolumn{1}{c}{{-0.38}} & 0.06                         \\ 
\textbf{3}                          & \multicolumn{1}{c}{1.24}                                  & \multicolumn{1}{c}{0.99}                         & 0.45               & \multicolumn{1}{c}{2.26}                         & \multicolumn{1}{c}{1.74}                         & 0.81                         \\ 
\textbf{4}                          & \multicolumn{1}{c}{2.22}                                  & \multicolumn{1}{c}{1.88}                         & 1.35               & \multicolumn{1}{c}{{-0.44}} & \multicolumn{1}{c}{{-0.41}} & 0.01                         \\ 
\textbf{5}                          & \multicolumn{1}{c}{0.68}                                  & \multicolumn{1}{c}{0.55}                         & 0.64               & \multicolumn{1}{c}{0.92}                         & \multicolumn{1}{c}{0.72}                         & 0.39                         \\ \hline 
\multicolumn{1}{l|}{\textbf{Avg. of 5}} & \multicolumn{1}{c}{\textbf{0.76}}                                  & \multicolumn{1}{c}{\textbf{0.63}}                         & \textbf{0.66}               & \multicolumn{1}{c}{\textbf{0.36}}                         & \multicolumn{1}{c}{\textbf{0.25}}                         & \textbf{0.24}                         \\ \hline
\multicolumn{1}{l|}{\textbf{Avg. of 3}} & \multicolumn{1}{c}{\textbf{1.27}}                                  & \multicolumn{1}{c}{\textbf{1.03}}                         & \textbf{0.91}               & \multicolumn{1}{c}{\textbf{0.02}}                         & \multicolumn{1}{c}{\textbf{-0.03}}                         & \textbf{0.15}                         \\ \hline
\end{tabular}
\end{table}
To observe any difference caused by the contamination, we calculated the difference in segmentation quality between two adjacent contamination classes. This yields a delta for the segmentation of \textit{no contamination} and \textit{little contamination}, as well as of \textit{little contamination} and \textit{strong contamination}. To gain a global perspective, we averaged the differences of all splits. To further mitigate the impact of poorly distributed scenes between the splits, we also calculated the average results excluding the best and worst performing splits. As seen in Table \ref{results:quantitative:diff}, split 1 achieved the worst results with constant reduction in segmentation quality, while split 3 achieved the best improvement in segmentation quality. 
In total, the overall performance for contaminated objects is higher in congruence with the SAM results. We theorize, that the neural networks learned object shapes, like window frames, of transparent objects more efficiently than their other properties. The contamination could then facilitate a better transparency to background recognition, as the transparent objects become more opaque, which in turn triggers a higher focus on the transparent foreground, as seen in Scene 1 in \ref{sec:results:contamination_impact}.

\subsubsection{Qualitative Results}
\begin{figure}[!htbp]
    \centering
    \setlength{\tabcolsep}{.1em} 
\renewcommand{\arraystretch}{0.5}
    \begin{tabular}{ccccc}
\multirow{2}{*}{\textbf{}}  & \multicolumn{2}{c}{\makecell[c]{\textbf{Scene 1} \\ Improvement in Segmentation}}               & \multicolumn{2}{c}{\makecell[c]{\textbf{Scene 2} \\ Degradation in Segmentation}} \\
                            & \multicolumn{1}{c}{\textit{no cont.}} & \textit{strong cont.} & \multicolumn{1}{c}{\textit{no cont.}} & \textit{strong cont.} \\
        \raisebox{1.5cm}{\rotatebox[]{90}{RGB + seg.}} & \includegraphics[width=.238\linewidth]{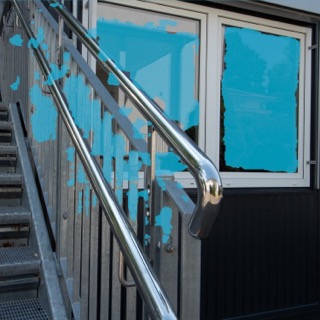} & \includegraphics[width=.238\linewidth]{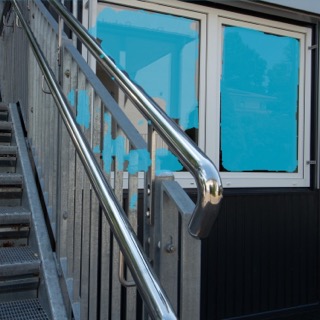} & \includegraphics[width=.238\linewidth]{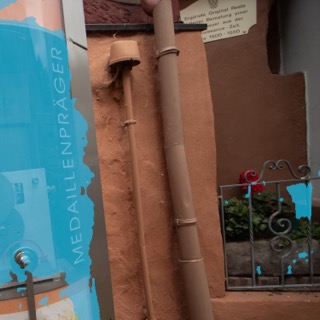} & \includegraphics[width=.238\linewidth]{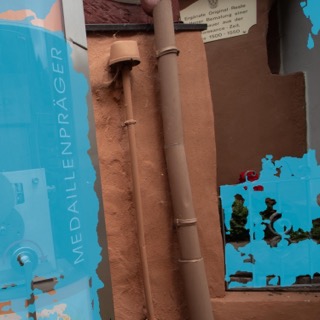} \\ 
        \raisebox{1.3cm}{\rotatebox[]{90}{segmentation}} & \includegraphics[width=.238\linewidth]{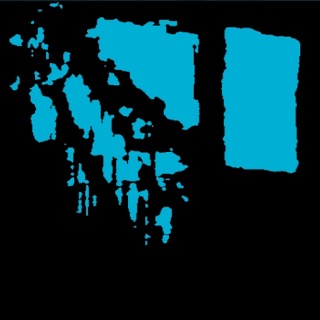} & \includegraphics[width=.238\linewidth]{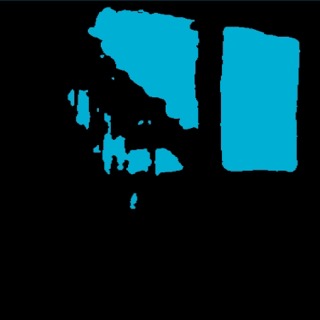} & \includegraphics[width=.238\linewidth]{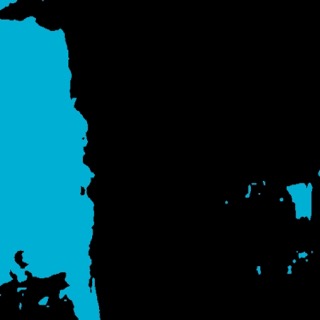} & \includegraphics[width=.238\linewidth]{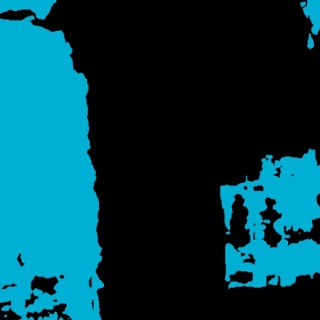} \\ 
        \raisebox{1.3cm}{\rotatebox[]{90}{ground truth}} & \includegraphics[width=.238\linewidth]{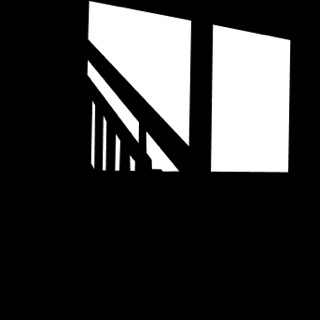} & \includegraphics[width=.238\linewidth]{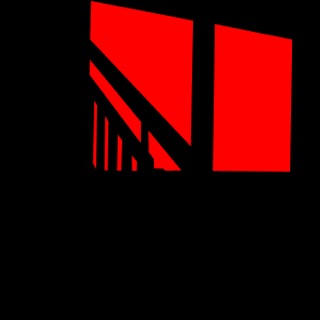} & \includegraphics[width=.238\linewidth]{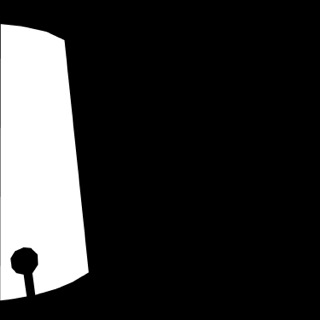} & \includegraphics[width=.238\linewidth]{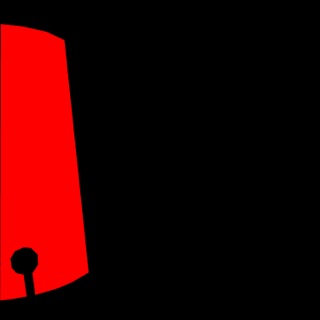} \\ 
        & 60.9 tIoU $\uparrow$ & 79.2 tIoU $\uparrow$ & 74.4 tIoU $\uparrow$ & 50.1 tIoU $\uparrow$ \\ 
    \end{tabular}
    \caption{Example scenes where the application of contamination could influence the segmentation quality in a meaningful way. The level of contamination is encoded through the colors of the ground truth annotations, with white denoting \textit{no contamination}, while red denotes \textit{strong contamination}. For \textbf{Scene 1}, presence of contamination did improve the segmentation, whereas for \textbf{Scene 2}, our simulation degraded the segmentation quality. This happens most likely through the fence structure, which is very similar to a structure that can encapsulate a glass pane.}
    \label{fig:results:qualitative_impact}
\end{figure}

This section highlights two scenes for a better understanding of the possible difference in segmentation caused by the contamination. In Fig. \ref{fig:results:qualitative_impact}, the segmentation performance of \textit{scene 1} is increased by the presence of our contamination, resulting in a gain of 18.6 \% tIoU comparing the results of \textit{no contamination} with \textit{strong contamination}. \textit{Scene 2} serves as an example in which the application of the contamination led to a decline of 24.3 \% tIoU. In this case, the misclassification happens most likely through the fence structure, which is very similar to a structure that can encapsulate a glass pane. A relatable behaviour of filling out possible structure is seen in Scene 1, where the border of the window are detected more accurately. We suppose, that the network learns the shapes of possible glass pane holders and fails to distinguish between transparency and no transparency in ambiguous or very hard cases. This should be solvable through more suitable data for fringe cases.

\subsection{Grade of Contamination Detection}
As an additional task, we evaluate the segmentation performance for the three levels of contamination. 
We achieve this by repeating the transfer-training process with four instead of two classes, as the annotations of our dataset encode the type of applied contamination.
This way, the model learns to segment the classes \textit{background, no contamination, little contamination} and \textit{strong contamination}.

\subsubsection{Quantitative Results}
As can be observed in Table \ref{table:results:dirt_detection}, the model did not match the quality of the transparency segmentation observed in Section \ref{sec:results:contamination_impact}. Nevertheless, it was still able to distinct between the different classes of contamination, most notably between \textit{no contamination} and \textit{strong contamination}, reaching results of up to 54.54 \% no cont. IoU and 48.18 \% strong cont. IoU. 

\begin{table}[!htbp]
\centering
\newcolumntype{P}{>{\centering\arraybackslash}p}
\newcolumntype{C}{>{\centering\arraybackslash}X}
\renewcommand\tabularxcolumn[1]{m{#1}}
\caption{Results after training the \textit{Trans4Trans} model to segment the different types of contamination. No cont., little cont. and strong cont. denotes respective contamination class, i.e. \textit{no contamination, little contamination} and \textit{strong contamination}.}
\label{table:results:dirt_detection}
\begin{tabularx}{.8\linewidth}{c|>{\hsize=0.33\hsize}C >{\hsize=0.33\hsize}C >{\hsize=0.4\hsize}C >{\hsize=0.33\hsize}C >{\hsize=0.33\hsize}C >{\hsize=0.33\hsize}C}
\hline
\textbf{Splits}                     & mIoU $\uparrow$ & PAcc $\uparrow$ & background \newline IoU $\uparrow$ & no \newline cont. \newline IoU $\uparrow$ & little \newline cont. \newline IoU $\uparrow$ & strong \newline cont. \newline IoU $\uparrow$ \\ \hline
\textbf{1}                          & 53.69 & 82.44 & 90.61          & 51.68        & 30.27            & 42.19            \\
\textbf{2}                          & 50.43  & 80.04 & \textbf{93.40}          & 43.81        & 21.40             & 43.10            \\
\textbf{3}                          & 54.21  & 82.62 & 90.61           & \textbf{54.54}         & 28.69             & 43.01             \\ 
\textbf{4}                          & 53.46  & 79.98 & 90.14          & 44.47        & 31.07             & \textbf{48.18}            \\
\textbf{5}                          & \textbf{55.64}  & \textbf{82.89} & 90.41           & 50.90         & \textbf{35.51}             & 45.77             \\ \hline
\multicolumn{1}{l|}{\textbf{Avg.}} & \textbf{53.49}  & \textbf{81.59} & \textbf{91.03}           & \textbf{49.08}         & \textbf{29.39}             & \textbf{44.45}             \\ \hline
\multicolumn{1}{l|}{\textbf{$\sigma$}} & \textbf{1.71}  & \textbf{1.30} & \textbf{1.19}           & \textbf{4.22}         & \textbf{4.59}             & \textbf{2.22}             \\ \hline
\end{tabularx}
\end{table}

\subsubsection{Qualitative Results}
\begin{figure}[!htbp]
    \centering
    \setlength{\tabcolsep}{.1em} 
\renewcommand{\arraystretch}{0.5}
    \begin{tabular}{ccccccc}
\multirow{3}{*}{\textbf{}}  & \multicolumn{3}{c}{\textbf{Scene 1}}  & \multicolumn{3}{c}{\textbf{Scene 2}} \\
    & \multicolumn{1}{c}{\textit{no cont.}} & \textit{little cont.} & \textit{strong cont.} & \multicolumn{1}{c}{\textit{no cont.}} & \textit{little cont.} & \textit{strong cont.} \\
        \raisebox{.8cm}{\rotatebox[]{90}{RGB}} & \includegraphics[width=.15\linewidth]{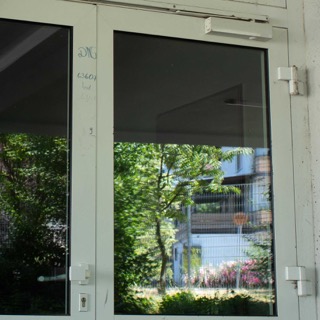} & \includegraphics[width=.15\linewidth]{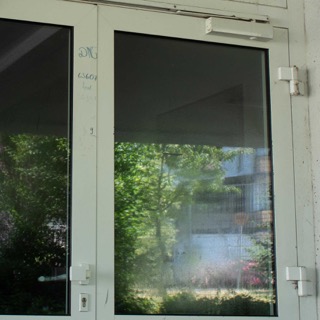} & \includegraphics[width=.15\linewidth]{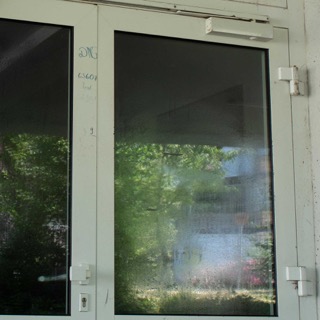} & \includegraphics[width=.15\linewidth]{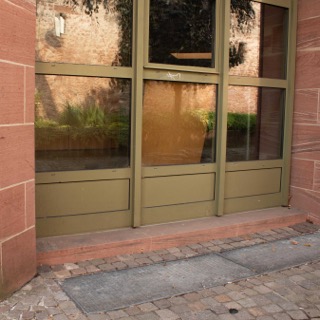} & \includegraphics[width=.15\linewidth]{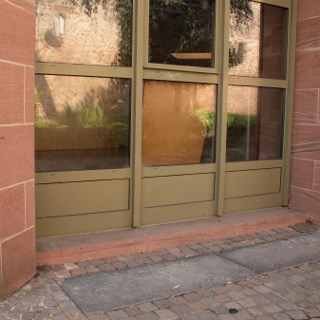} & \includegraphics[width=.15\linewidth]{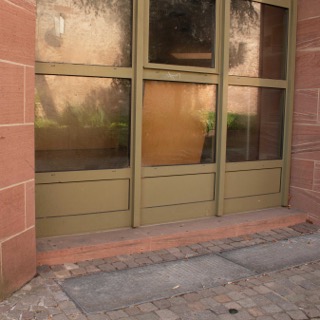} \\ 
        \raisebox{1cm}{\rotatebox[]{90}{segmen.}} & \includegraphics[width=.15\linewidth]{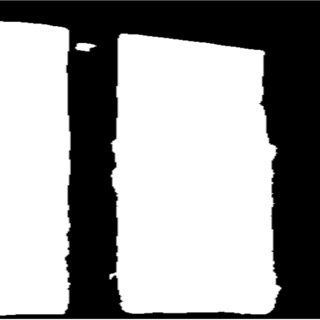} & \includegraphics[width=.15\linewidth]{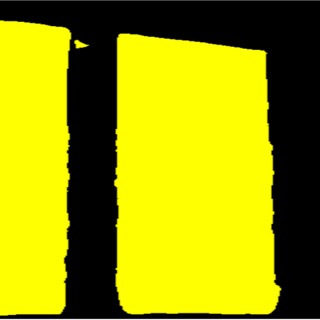} & \includegraphics[width=.15\linewidth]{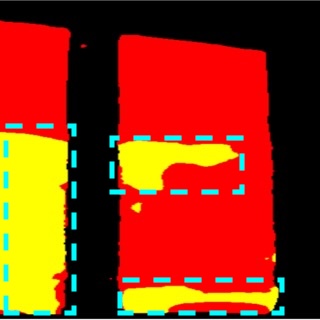} & \includegraphics[width=.15\linewidth]{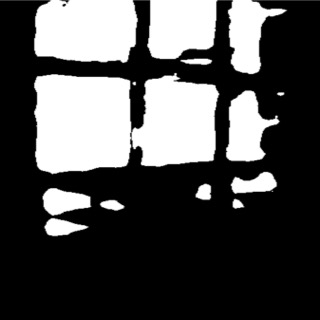} & \includegraphics[width=.15\linewidth]{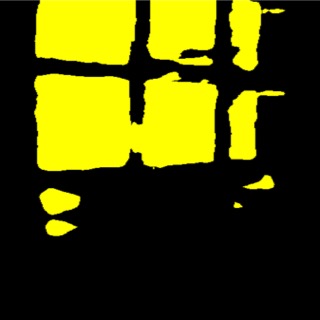} & \includegraphics[width=.15\linewidth]{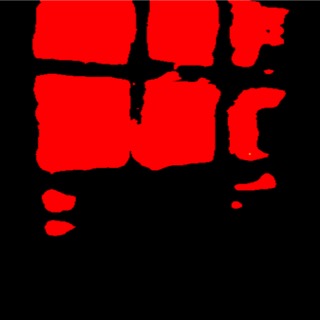} \\ 
        \raisebox{0.75cm}{\rotatebox[]{90}{GT}} & \includegraphics[width=.15\linewidth]{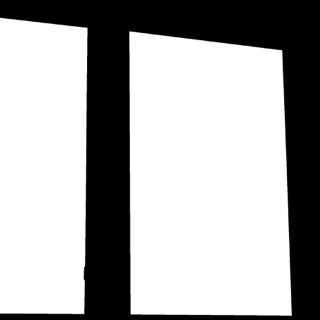} & \includegraphics[width=.15\linewidth]{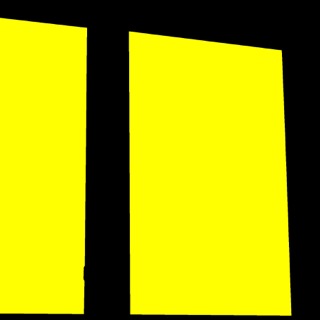} & \includegraphics[width=.15\linewidth]{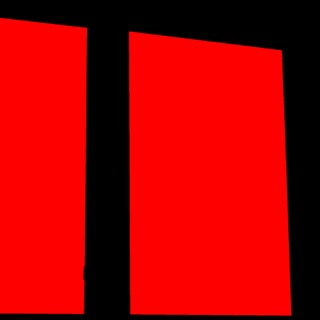} & \includegraphics[width=.15\linewidth]{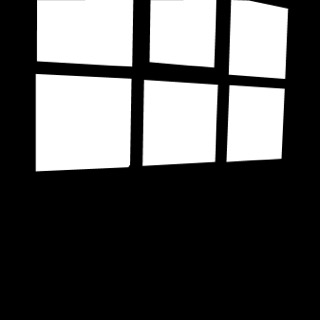} & \includegraphics[width=.15\linewidth]{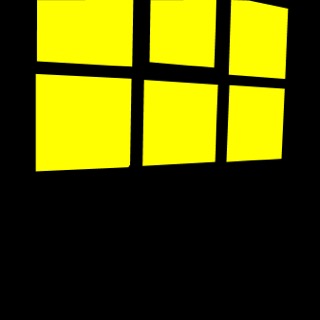} & \includegraphics[width=.15\linewidth]{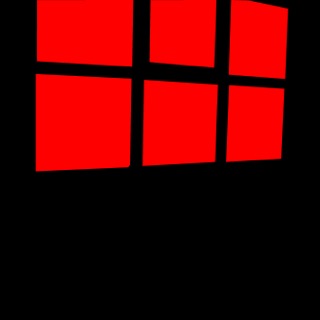} \\ 
    \end{tabular}
    \caption{Examples for scenes in which the model was able to properly segment the type of contamination applied to the surface. A white segmentation denotes the prediction of \textit{no contamination}, yellow the prediction of \textit{little contamination} and red the detection of \textit{strong contamination}. In \textbf{Scene 1}, the model failed to fully distinguish between little and strong contamination in the marked areas.}
    \label{fig:results:qualitative_dirt_detection}
\end{figure}
To better visualize the segmentation of different types of contamination, we selected a set of example scenes to demonstrate the capabilities of the \textit{Trans4Trans} model for this task. Fig. \ref{fig:results:qualitative_dirt_detection} illustrates, that the model was able to properly detect the type of contamination present on the surfaces for the displayed scenes. For \textbf{Scene 1}, the model mistook the highlighted areas for \textit{little contamination}, although the ground-truth information for this image was \textit{strong contamination}. When analyzing the RGB images, the marked areas show little difference to the image with \textit{little contamination}, which indicates that the model tries to localize the type of contamination.

\section{Conclusion and Future Work}
Transparent structures, such as protective glass in industrial areas, are under the influence of contamination like every other object of our world. However, the resulting change in appearance and the direct influence on objects behind the transparency are more prominent. We propose a novel dataset with 489 images and three categories of contamination to assess changes in transparency segmentation and contamination categorization. Our evaluation is based on the \textit{Trans4Trans} \cite{zhang2021trans4trans} model and the \textit{Trans10K} dataset as additional training data. Our results show, that transparency segmentation capabilities improve due to contamination. Furthermore, the different levels of contamination are distinguishable during segmentation, albeit at a lower quality due to the more complex task and increased segmentation classes. Our findings therefore suggest, that it is not only easier to find contaminated transparent objects, but also to determine whether they should be cleaned soon. This is especially useful to provide resilience against anomalies or shifts, when a vision system is behind contamination prone protection glass. In the future, we want to combine our work with additional insights about transparent objects and apply it to real-world use cases. This should yield additional insights to improve the reliability of computer vision applications in industrial environments. Furthermore, we want to improve the performance through more sophisticated data, which encompasses the hard cases we determined in this work.

\begin{credits}
\end{credits}
%
%
%
\bibliographystyle{splncs04}
\bibliography{references}

\begin{thebibliography}{10}
\providecommand{\url}[1]{\texttt{#1}}
\providecommand{\urlprefix}{URL }
\providecommand{\doi}[1]{https://doi.org/#1}

\bibitem{badrinarayanan2017segnet}
Badrinarayanan, V., Kendall, A., Cipolla, R.: Segnet: A deep convolutional
  encoder-decoder architecture for image segmentation. IEEE transactions on
  pattern analysis and machine intelligence  \textbf{39}(12),  2481--2495
  (2017)

\bibitem{bormann2020dirtnet}
Bormann, R., Wang, X., Xu, J., Schmidt, J.: Dirtnet: Visual dirt detection for
  autonomous cleaning robots. In: 2020 IEEE International Conference on
  Robotics and Automation (ICRA). pp. 1977--1983. IEEE (2020)

\bibitem{canedo2021deep}
Canedo, D., Fonseca, P., Georgieva, P., Neves, A.J.: A deep learning-based dirt
  detection computer vision system for floor-cleaning robots with improved data
  collection. Technologies  \textbf{9}(4), ~94 (2021)

\bibitem{chen2018tom}
Chen, G., Han, K., Wong, K.Y.K.: Tom-net: Learning transparent object matting
  from a single image. In: IEEE/CVF CVPR conference proceedings (2018)

\bibitem{chen2014semantic}
Chen, L.C., Papandreou, G., Kokkinos, I., Murphy, K., Yuille, A.L.: Semantic
  image segmentation with deep convolutional nets and fully connected crfs.
  arXiv preprint arXiv:1412.7062  (2014)

\bibitem{chen2017deeplab}
Chen, L.C., Papandreou, G., Kokkinos, I., Murphy, K., Yuille, A.L.: Deeplab:
  Semantic image segmentation with deep convolutional nets, atrous convolution,
  and fully connected crfs. IEEE transactions on pattern analysis and machine
  intelligence  \textbf{40}(4),  834--848 (2017)

\bibitem{chen2018encoder}
Chen, L.C., Zhu, Y., Papandreou, G., Schroff, F., Adam, H.: Encoder-decoder
  with atrous separable convolution for semantic image segmentation. In:
  Proceedings of the European conference on computer vision (ECCV). pp.
  801--818 (2018)

\bibitem{cheng2022masked}
Cheng, B., Misra, I., Schwing, A.G., Kirillov, A., Girdhar, R.:
  Masked-attention mask transformer for universal image segmentation. In:
  IEEE/CVF CVPR conference proceedings. pp. 1290--1299 (2022)

\bibitem{chu2021twins}
Chu, X., Tian, Z., Wang, Y., Zhang, B., Ren, H., Wei, X., Xia, H., Shen, C.:
  Twins: Revisiting the design of spatial attention in vision transformers.
  Advances in Neural Information Processing Systems  \textbf{34},  9355--9366
  (2021)

\bibitem{chu2021conditional}
Chu, X., Tian, Z., Zhang, B., Wang, X., Wei, X., Xia, H., Shen, C.: Conditional
  positional encodings for vision transformers. arXiv preprint arXiv:2102.10882
   (2021)

\bibitem{dosovitskiy2020image}
Dosovitskiy, A., Beyer, L., Kolesnikov, A., Weissenborn, D., Zhai, X.,
  Unterthiner, T., Dehghani, M., Minderer, M., Heigold, G., Gelly, S., et~al.:
  An image is worth 16x16 words: Transformers for image recognition at scale.
  preprint arXiv:2010.11929  (2020)

\bibitem{eigen2013restoring}
Eigen, D., Krishnan, D., Fergus, R.: Restoring an image taken through a window
  covered with dirt or rain. In: IEEE ICCV proceedings. pp. 633--640 (2013)

\bibitem{fang2023eva}
Fang, Y., Wang, W., Xie, B., Sun, Q., Wu, L., Wang, X., Huang, T., Wang, X.,
  Cao, Y.: Eva: Exploring the limits of masked visual representation learning
  at scale. In: IEEE/CVF CVPR conference proceedings. pp. 19358--19369 (2023)

\bibitem{halimeh2009raindrop}
Halimeh, J.C., Roser, M.: Raindrop detection on car windshields using
  geometric-photometric environment construction and intensity-based
  correlation. In: 2009 IEEE Intelligent Vehicles Symposium. pp. 610--615. IEEE
  (2009)

\bibitem{he2015spatial}
He, K., Zhang, X., Ren, S., Sun, J.: Spatial pyramid pooling in deep
  convolutional networks for visual recognition. IEEE transactions on pattern
  analysis and machine intelligence  \textbf{37}(9),  1904--1916 (2015)

\bibitem{he2016deep}
He, K., Zhang, X., Ren, S., Sun, J.: Deep residual learning for image
  recognition. In: Proceedings of the IEEE conference on computer vision and
  pattern recognition. pp. 770--778 (2016)

\bibitem{huo2023glass}
Huo, D., Wang, J., Qian, Y., Yang, Y.H.: Glass segmentation with rgb-thermal
  image pairs. IEEE Transactions on Image Processing  \textbf{32},  1911--1926
  (2023)

\bibitem{iglovikov2018ternausnetv2}
Iglovikov, V., Seferbekov, S., Buslaev, A., Shvets, A.: Ternausnetv2: Fully
  convolutional network for instance segmentation. In: IEEE/CVF CVPR conference
  proceedings workshops. pp. 233--237 (2018)

\bibitem{jain2023oneformer}
Jain, J., Li, J., Chiu, M.T., Hassani, A., Orlov, N., Shi, H.: Oneformer: One
  transformer to rule universal image segmentation. In: IEEE/CVF CVPR
  conference proceedings. pp. 2989--2998 (2023)

\bibitem{jimenez2019dirt}
Jim{\'e}nez, A.A., Mu{\~n}oz, C.Q.G., M{\'a}rquez, F.P.G.: Dirt and mud
  detection and diagnosis on a wind turbine blade employing guided waves and
  supervised learning classifiers. Reliability Engineering \& System Safety
  \textbf{184},  2--12 (2019)

\bibitem{kirillov2023segment}
Kirillov, A., Mintun, E., Ravi, N., Mao, H., Rolland, C., Gustafson, L., Xiao,
  T., Whitehead, S., Berg, A.C., Lo, W.Y., et~al.: Segment anything. arXiv
  preprint arXiv:2304.02643  (2023)

\bibitem{knauthe2023distortion}
Knauthe, V., P{\"o}llabauer, T., Faller, K., Kraus, M., Wirth, T., Buelow,
  M.v., Kuijper, A., Fellner, D.W.: Distortion-based transparency detection
  using deep learning on a novel synthetic image dataset. In: Scandinavian
  Conference on Image Analysis. pp. 251--267. Springer (2023)

\bibitem{li2018pyramid}
Li, H., Xiong, P., An, J., Wang, L.: Pyramid attention network for semantic
  segmentation. arXiv preprint arXiv:1805.10180  (2018)

\bibitem{lin2016efficient}
Lin, G., Shen, C., Van Den~Hengel, A., Reid, I.: Efficient piecewise training
  of deep structured models for semantic segmentation. In: IEEE/CVF CVPR
  conference proceedings. pp. 3194--3203 (2016)

\bibitem{liu2021swin}
Liu, Z., Lin, Y., Cao, Y., Hu, H., Wei, Y., Zhang, Z., Lin, S., Guo, B.: Swin
  transformer: Hierarchical vision transformer using shifted windows. In:
  IEEE/CVF ICCV conference proceedings. pp. 10012--10022 (2021)

\bibitem{long2015fully}
Long, J., Shelhamer, E., Darrell, T.: Fully convolutional networks for semantic
  segmentation. In: IEEE/CVF CVPR conference proceedings. pp. 3431--3440 (2015)

\bibitem{nguyen2022detection}
Nguyen, H.T., Tsao, Y.M., Wang, H.C.: Detection of weak micro-scratches on
  aspherical lenses using a gabor neural network and transfer learning. Applied
  Optics  \textbf{61}(20),  6046--6056 (2022)

\bibitem{olorunfemi2022solar}
Olorunfemi, B.O., Ogbolumani, O.A., Nwulu, N.: Solar panels dirt monitoring and
  cleaning for performance improvement: a systematic review on smart systems.
  Sustainability  \textbf{14}(17),  10920 (2022)

\bibitem{oquab2023dinov2}
Oquab, M., Darcet, T., Moutakanni, T., Vo, H., Szafraniec, M., Khalidov, V.,
  Fernandez, P., Haziza, D., Massa, F., El-Nouby, A., et~al.: Dinov2: Learning
  robust visual features without supervision. arXiv preprint arXiv:2304.07193
  (2023)

\bibitem{quan2019deep}
Quan, Y., Deng, S., Chen, Y., Ji, H.: Deep learning for seeing through window
  with raindrops. In: IEEE/CVF ICCV conference proceedings. pp. 2463--2471
  (2019)

\bibitem{ronneberger2015u}
Ronneberger, O., Fischer, P., Brox, T.: U-net: Convolutional networks for
  biomedical image segmentation. In: Medical Image Computing and
  Computer-Assisted Intervention--MICCAI 2015: 18th International Conference,
  Munich, Germany, October 5-9, 2015, Proceedings, Part III 18. pp. 234--241.
  Springer (2015)

\bibitem{shajahan2021camera}
Shajahan, J.M.A., Reyes, S.M., Xiao, J.: Camera lens dust detection and dust
  removal for mobile robots in dusty fields. In: 2021 IEEE International
  Conference on Robotics and Biomimetics (ROBIO). pp. 687--691. IEEE (2021)

\bibitem{shelhamer2017fully}
Shelhamer, E., Long, J., Darrell, T., et~al.: Fully convolutional networks for
  semantic segmentation. IEEE Trans. Pattern Anal. Mach. Intell.
  \textbf{39}(4),  640--651 (2017)

\bibitem{strudel2021segmenter}
Strudel, R., Garcia, R., Laptev, I., Schmid, C.: Segmenter: Transformer for
  semantic segmentation. In: IEEE/CVF ICCV proceedings. pp. 7262--7272 (2021)

\bibitem{su2023towards}
Su, W., Zhu, X., Tao, C., Lu, L., Li, B., Huang, G., Qiao, Y., Wang, X., Zhou,
  J., Dai, J.: Towards all-in-one pre-training via maximizing multi-modal
  mutual information. In: IEEE/CVF CVPR conference proceedings. pp.
  15888--15899 (2023)

\bibitem{thisanke2023semantic}
Thisanke, H., Deshan, C., Chamith, K., Seneviratne, S., Vidanaarachchi, R.,
  Herath, D.: Semantic segmentation using vision transformers: A survey.
  Engineering Applications of Artificial Intelligence  \textbf{126},  106669
  (2023)

\bibitem{ulku2022survey}
Ulku, I., Akag{\"u}nd{\"u}z, E.: A survey on deep learning-based architectures
  for semantic segmentation on 2d images. Applied Artificial Intelligence
  \textbf{36}(1),  2032924 (2022)

\bibitem{uricar2021let}
Uricar, M., Sistu, G., Rashed, H., Vobecky, A., Kumar, V.R., Krizek, P.,
  Burger, F., Yogamani, S.: Let's get dirty: Gan based data augmentation for
  camera lens soiling detection in autonomous driving. In: Proceedings of the
  IEEE/CVF winter conference on applications of computer vision. pp. 766--775
  (2021)

\bibitem{vaswani2017attention}
Vaswani, A., Shazeer, N., Parmar, N., Uszkoreit, J., Jones, L., Gomez, A.N.,
  Kaiser, {\L}., Polosukhin, I.: Attention is all you need. Advances in neural
  information processing systems  \textbf{30} (2017)

\bibitem{wang2023one}
Wang, P., Wang, S., Lin, J., Bai, S., Zhou, X., Zhou, J., Wang, X., Zhou, C.:
  One-peace: Exploring one general representation model toward unlimited
  modalities. arXiv preprint arXiv:2305.11172  (2023)

\bibitem{wang2023internimage}
Wang, W., Dai, J., Chen, Z., Huang, Z., Li, Z., Zhu, X., Hu, X., Lu, T., Lu,
  L., Li, H., et~al.: Internimage: Exploring large-scale vision foundation
  models with deformable convolutions. In: IEEE/CVF CVPR proceedings. pp.
  14408--14419 (2023)

\bibitem{wang2021pyramid}
Wang, W., Xie, E., Li, X., Fan, D.P., Song, K., Liang, D., Lu, T., Luo, P.,
  Shao, L.: Pyramid vision transformer: A versatile backbone for dense
  prediction without convolutions. In: IEEE/CVF ICCV conference proceedings.
  pp. 568--578 (2021)

\bibitem{wang2022pvt}
Wang, W., Xie, E., Li, X., Fan, D.P., Song, K., Liang, D., Lu, T., Luo, P.,
  Shao, L.: Pvt v2: Improved baselines with pyramid vision transformer.
  Computational Visual Media  \textbf{8}(3),  415--424 (2022)

\bibitem{wang2022image}
Wang, W., Bao, H., Dong, L., Bjorck, J., Peng, Z., Liu, Q., Aggarwal, K.,
  Mohammed, O.K., Singhal, S., Som, S., et~al.: Image as a foreign language:
  Beit pretraining for all vision and vision-language tasks. preprint
  arXiv:2208.10442  (2022)

\bibitem{xie2021segformer}
Xie, E., Wang, W., Yu, Z., Anandkumar, A., Alvarez, J.M., Luo, P.: Segformer:
  Simple and efficient design for semantic segmentation with transformers.
  Advances in Neural Information Processing Systems  \textbf{34},  12077--12090
  (2021)

\bibitem{xie2020segmenting}
Xie, E., Wang, W., Wang, W., Ding, M., Shen, C., Luo, P.: Segmenting
  transparent objects in the wild. In: Computer Vision--ECCV 2020: 16th
  European Conference, Glasgow, UK, Proceedings, Part XIII 16. pp. 696--711.
  Springer (2020)

\bibitem{xie2021segmenting}
Xie, E., Wang, W., Wang, W., Sun, P., Xu, H., Liang, D., Luo, P.: Segmenting
  transparent object in the wild with transformer. preprint arXiv:2101.08461
  (2021)

\bibitem{xu2015transcut}
Xu, Y., Nagahara, H., Shimada, A., Taniguchi, R.i.: Transcut: Transparent
  object segmentation from a light-field image. In: IEEE/CVF ICCV conference
  proceedings. pp. 3442--3450 (2015)

\bibitem{yu2018bisenet}
Yu, C., Wang, J., Peng, C., Gao, C., Yu, G., Sang, N.: Bisenet: Bilateral
  segmentation network for real-time semantic segmentation. In: Proceedings of
  the European conference on computer vision (ECCV). pp. 325--341 (2018)

\bibitem{yuan2021hrformer}
Yuan, Y., Fu, R., Huang, L., Lin, W., Zhang, C., Chen, X., Wang, J.: Hrformer:
  High-resolution vision transformer for dense predict. Advances in Neural
  Information Processing Systems  \textbf{34},  7281--7293 (2021)

\bibitem{zhang2021trans4trans}
Zhang, J., Yang, K., Constantinescu, A., Peng, K., M{\"u}ller, K.,
  Stiefelhagen, R.: Trans4trans: Efficient transformer for transparent object
  segmentation to help visually impaired people navigate in the real world. In:
  IEEE/CVF ICCV conference proceedings. pp. 1760--1770 (2021)

\bibitem{zhang2022trans4trans}
Zhang, J., Yang, K., Constantinescu, A., Peng, K., M{\"u}ller, K.,
  Stiefelhagen, R.: Trans4trans: Efficient transformer for transparent object
  and semantic scene segmentation in real-world navigation assistance. IEEE
  Transactions on Intelligent Transportation Systems  \textbf{23}(10),
  19173--19186 (2022)

\bibitem{zhao2017pyramid}
Zhao, H., Shi, J., Qi, X., Wang, X., Jia, J.: Pyramid scene parsing network.
  In: IEEE/CVF CVPR conference proceedings. pp. 2881--2890 (2017)

\bibitem{zheng2015conditional}
Zheng, S., Jayasumana, S., Romera-Paredes, B., Vineet, V., Su, Z., Du, D.,
  Huang, C., Torr, P.H.: Conditional random fields as recurrent neural
  networks. In: IEEE/CVF ICCV conference proceedings. pp. 1529--1537 (2015)

\bibitem{zheng2021rethinking}
Zheng, S., Lu, J., Zhao, H., Zhu, X., Luo, Z., Wang, Y., Fu, Y., Feng, J.,
  Xiang, T., Torr, P.H., et~al.: Rethinking semantic segmentation from a
  sequence-to-sequence perspective with transformers. In: IEEE/CVF CVPR
  conference proceedings. pp. 6881--6890 (2021)

\end{thebibliography}
%





\end{document}